\documentclass{article}

\usepackage[preprint]{neurips_2026}

\usepackage{xspace}
\usepackage{amsmath}
\usepackage{amssymb}

\usepackage{multirow}
\usepackage{makecell}
\usepackage{wrapfig}
\usepackage{colortbl}

\usepackage[utf8]{inputenc} % allow utf-8 input
\usepackage[T1]{fontenc}    % use 8-bit T1 fonts
\usepackage{hyperref}       % hyperlinks
\usepackage{url}            % simple URL typesetting
\usepackage{booktabs}       % professional-quality tables
\usepackage{amsfonts}       % blackboard math symbols
\usepackage{nicefrac}       % compact symbols for 1/2, etc.
\usepackage{microtype}      % microtypography
\usepackage{xcolor}         % colors
\usepackage[colorinlistoftodos,prependcaption,textsize=tiny]{todonotes}
\usepackage[table]{xcolor}

\definecolor{lm_purple_low}{RGB}{240,240,248}
\definecolor{lm_purple}{RGB}{227,227,240}
\definecolor{lm_red}{RGB}{230,36,43}

\title{\methodname: Retrieval-Based Synthesis of Interoperable Multi-Agent Workflows}

\author{%
  \textbf{Shuaike Shen}$^{1, *}$, \textbf{Wenduo Cheng}$^{1, *}$, \textbf{Shike Wang}$^1$, \textbf{Mingqian Ma}$^2$, \textbf{Jian Ma}$^{1\dagger}$\\ 
$^1$Ray and Stephanie Lane Computational Biology Department,\\ School of Computer Science, Carnegie Mellon University
\\$^2$Machine Learning Department,\\ School of Computer Science, Carnegie Mellon University\\
$^*$Equal contribution, $^\dagger$Correspondence: \texttt{jianma@cs.cmu.edu}\\
}

\begin{document}

\maketitle

\begin{abstract}
Designing multi-agent workflows is especially difficult in open-ended scientific settings where tasks lack curated training sets, reliable scalar evaluation metrics, and standardized interfaces between existing tools and agents.
We propose \methodname, a retrieval-based synthesis framework that composes reusable skills, tools, and external agents into executable workflows through typed artifact handoffs, then applies bounded self-guided local repair to implicated components when execution evidence indicates failure. 
In two open-world genomics case studies, \methodname composes independently developed scientific agents and external tool repositories into auditable workflows without redesigning them or running global topology search. 
It coordinates specialized agents for spatial transcriptomics and gene-set interpretation to enable collaborative discovery from spatial transcriptomics data, and builds a parallel workflow for cross-modality marker analysis on single-cell multiome data.
\methodname can also import a searched workflow as a structural prior and improve it by grounding nodes with retrieved components and applying local repair, showing that synthesis and search are complementary. 
On six coding, math, and question-answering benchmarks, \methodname achieves the best result on four benchmarks and the best average score under a unified backbone setting, while consistently reducing per-task cost relative to multi-agent baselines.
Together, these results suggest that retrieval-based synthesis can extend automated agentic workflow design beyond benchmark-optimized agent graphs to open-world workflows built from existing agents, tools, and typed artifacts.
Code and project website s are available at \url{https://github.com/ma-compbio-lab/AgentCo-Op} and \url{https://ma-compbio-lab.github.io/AgentCo-Op/}
\end{abstract}

\section{Introduction}
\label{sec:introduction}
Multi-agent LLM systems decompose complex tasks across specialized roles, tools, and prompts, and have shown strong results on reasoning, coding, question answering (QA), and scientific analysis~\citep{tran2025multi, hong2023metagpt, wu2023autogen}. As these systems mature, the bottleneck has shifted from constructing individual agents to designing interoperable workflows among them. Recent automated methods such as ADAS~\citep{hu2024adas}, AFlow~\citep{zhang2024aflow}, and AgentSquare~\citep{shang2024agentsquare} frame this design problem as searching over candidate topologies, prompts, operators, or workflow programs, optimizing against a training set with a scalar evaluation function. This formulation is powerful when representative tasks and reliable scalar signals are available, and has shown strong results on standard QA, math, and coding benchmarks.

However, this search-based formulation becomes limiting for a broad class of real-world tasks. In scientific domains, problems are often open-ended and rarely come with curated training sets, standardized test cases, or automatic evaluation functions that reflect scientific utility. In genomics, for example, marker-gene interpretation through pathway or gene-set enrichment has no single ground-truth answer; the same gene list can support multiple plausible interpretations depending on tissue, cell type, disease context, database choice, and statistical threshold~\citep{subramanian2005gene,wang2025geneagent}. Such tasks are judged through heterogeneous intermediate evidence such as statistical significance, biological plausibility, consistency with known markers, and provenance of the analysis, which are hard to compress into a single reward, making repeated scoring of candidate workflows expensive and often impractical.

A second challenge concerns interoperability rather than optimization. Many scientific domains already have tool-augmented agents that experts have built and validated for specialized tasks, so a substantial part of the challenge is coordinating independently developed systems rather than creating new capabilities from scratch~\citep{wei2025ai}. Such agents typically rely on incompatible environments, expose different interfaces, and maintain separate provenance states, so simply placing multiple agents together does not yield a coherent workflow. What is needed is a mechanism for retrieving relevant components, aligning their interfaces, passing typed artifacts between them, and repairing failed components using execution evidence.

We propose \methodname, a framework that reframes automated multi-agent workflow design as retrieval-based synthesis. Given a task specification, \methodname retrieves relevant resources, skills, tools, and external agents from curated libraries or user-provided repositories, assigns them to specialized roles, aligns their input-output interfaces through typed artifacts, and synthesizes them into an executable workflow as a directed graph. During execution, \methodname monitors heterogeneous evidence such as execution traces, validation checks, tool errors, and cost signals, and triggers bounded evidence-guided local repair on implicated components rather than restarting synthesis. This synthesis-first view produces workflows where scalar metrics are unavailable, reuses prior engineering effort by composing existing skills, tools, and entire repositories of independently developed agents, and confines local repair to failing components rather than repeating global search.

We evaluate \methodname in both open-world scientific settings and standard benchmarks. The open-world setting motivates the synthesis-first design, since benchmark-driven search is often impractical when no curated benchmark or evaluation function is available. We study three representative applications. First, \methodname coordinates independently developed domain agents through a serial repository handoff, composing TissueAgent and GeneAgent for differential expression and gene-set interpretation on a developing human heart MERFISH dataset. Second, \methodname composes complementary domain workflows in parallel, integrating Seurat and Signac into a cross-modality marker-discovery pipeline on PBMC multiome data. Third, \methodname reuses existing agent graphs by importing prior workflows, grounding their nodes with retrieved skills and tools, and applying bounded local repair during execution. On six standard QA, mathematical reasoning, and code generation benchmarks, \methodname further achieves the best performance on four of the six benchmarks under a matched backbone setting and the lowest average cost.

Our contributions are as follows.
\vspace{-0.8em}
\begin{enumerate}
    \item We formulate automated multi-agent workflow design as retrieval-based synthesis for settings where scalar rewards are weak or unavailable, and instantiate this view in \methodname, a framework that dynamically composes resources, skills, tools, and external agents into executable workflows through typed artifact handoffs and bounded evidence-guided local repair.
\vspace{-0.3em}

    \item We demonstrate that \methodname can coordinate independently developed scientific agents and tool repositories in open-world genomics tasks. Given only a task specification and GitHub links to relevant repositories, \methodname automatically synthesizes interoperable multi-agent workflows that support collaboration among heterogeneous methods.
    % We demonstrate open-world scientific workflow composition through two genomics case studies, coordinating TissueAgent with GeneAgent in a serial handoff and Seurat with Signac in a parallel cross-modality pipeline through validated typed artifacts.
\vspace{-0.3em}

    \item We further show that synthesis and search are complementary by importing an AFlow-searched agentic workflow on MBPP and improving it through retrieval grounding and evidence-guided local repair.
\vspace{-0.3em}

    \item On six coding, math, and QA benchmarks, \methodname is competitive with search-based agentic workflow design methods, achieving the best performance on four of six benchmarks while consistently lowering test-time token cost.
\vspace{-0.3em}
    
\end{enumerate}
\vspace{-1em}

\section{Related Work}
\label{sec:related}
\vspace{-0.5em}
\subsection{Multi-agent systems}
\vspace{-0.3em}
Multi-agent LLM systems decompose tasks across agents with different roles, tools, and communication patterns. Role-based collaboration assigns agents complementary responsibilities, as in CAMEL~\citep{li2023camel}, MetaGPT~\citep{hong2023metagpt}, AutoGen~\citep{wu2023autogen}, and AgentVerse~\citep{chen2023agentverse}. Deliberation-based systems improve reasoning by having multiple agents propose, debate, or reconcile answers, as in LLM-Debate~\citep{du2024improving} and ReConcile~\citep{chen2024reconcile}. Practical guides further codify manager-style coordination, handoffs, guardrails, and subagents~\citep{openai2025guide,openai2025agentssdk,anthropic2025subagents}. These works treat agents as composable building blocks, but their workflow structures are still largely manually designed or template-based, which limits their generalization to new tasks.
\vspace{-1em}

\subsection{Automatic agentic workflow design}
\vspace{-0.2em}
A growing line of work automates the design of agentic workflows. Early systems such as DyLAN optimize team participation and communication through dynamic selection~\citep{liu2023dylan}, and GPTSwarm formulates agent collaboration as an optimizable graph~\citep{zhuge2024gptswarm}. More recent methods broaden the search space: ADAS searches over code-defined agents~\citep{hu2024adas}, AFlow uses Monte Carlo Tree Search over executable workflow graphs from execution feedback~\citep{zhang2024aflow}, AgentSquare defines a modular space over planning, reasoning, tool use, and memory~\citep{shang2024agentsquare}, and MaAS introduces an agentic supernet that samples query-dependent architectures~\citep{zhang2025multi}. Related efforts further explore automatic workflow generation and evolution, including Flow~\citep{niu2025flow}, EvoAgentX~\citep{wang2025evoagentx}, SEW~\citep{zhao2025sew}, and AutoFlow~\citep{li2024autoflow}. These methods typically rely on repeated proposal, execution, and evaluation under representative tasks and scalar feedback. \methodname targets a complementary setting where such feedback is weak, costly, or inaccessible, compiling a coordinated workflow directly from available skills, prior agents, and task requirements, while limiting runtime adaptation to bounded evidence-guided local repair.
\vspace{-1em}

\subsection{Agent skills and tool use}
\vspace{-0.2em}
A complementary line equips agents with externally specified capabilities. The Model Context Protocol standardizes tool, resource, and prompt access across providers~\citep{anthropic2024mcp}. Building on this, the Anthropic Agent Skills package procedural knowledge as portable folders loaded on demand via progressive disclosure~\citep{anthropic2025skills}, with a recent survey systematizing the paradigm~\citep{bhardwaj2026skillsurvey}. SkillFoundry mines heterogeneous resources into self-evolving skill libraries with executable contracts~\citep{shen2026skillfoundry}, and EvoSkills evolves multi-file skill packages through co-evolutionary verification~\citep{zhang2026evoskills}. Earlier tool-use work covers learned invocation and large API retrieval~\citep{schick2023toolformer, qin2023toolllm, patil2024gorilla}. These works expose capabilities to agents but do not determine how they should be organized into task-specific workflows. \methodname builds on this direction by treating skills as typed, testable units whose contracts are enforced during workflow synthesis and typed artifact handoff.
\vspace{-1em}

\subsection{Scientific agents}
\vspace{-0.2em}

LLM agents are increasingly applied to scientific discovery. SpatialAgent addresses spatial-biology pipelines from panel design to hypothesis generation~\citep{wang2025spatialagent}, and GeneAgent reduces hallucinations in gene-set analysis through database-grounded self-verification~\citep{wang2025geneagent}. The Virtual Lab orchestrates a principal investigator and specialist agents to design experimentally validated SARS-CoV-2 nanobodies~\citep{swanson2025virtuallab}. Biomni provides a generalist biomedical action space~\citep{huang2025biomni}, and STELLA self-evolves its template library and tool ocean~\citep{jin2025stella}. Other systems target gene editing, perturbation design, and chemistry, including CRISPR-GPT, BioDiscoveryAgent, and ChemCrow~\citep{huang2024crisprgpt,roohani2025biodiscoveryagent,boiko2023chemcrow}. These agents offer powerful specialized capabilities, but are typically built as standalone systems for specific task families. Composing them into multi-step, cross-modal, or interdisciplinary workflows remains difficult because their interfaces, environments, outputs, and assumptions are not aligned. \methodname addresses this composition problem by wrapping specialized agents and domain workflows as executable graph nodes, aligning them through typed artifacts, and synthesizing coherent collaborative workflows.
\vspace{-1em}
 
\section{Method}
\label{sec:method}
\vspace{-0.5em}
\subsection{Problem formulation}
\vspace{-0.5em}
We study the problem of automatically constructing multi-agent workflows for complex tasks. 
Given a task specification \(x\), the goal is to produce an executable workflow \(W\) that decomposes the task, grounds each role in retrieved components, and coordinates communication through typed artifacts. 
% and artifact passing among agents, and adapts the workflow based on execution evidence.
We represent a task specification as
% \vspace{-0.1em}
\begin{align}
    x = (g, c, r, \Omega),
\end{align}

\vspace{-0.5em}
where \(g\) is the user goal, \(c\) is the task context, \(r\) specifies operational constraints such as available data, budget, runtime, environment requirements, and desired output format, and \(\Omega\) denotes task-specific resources provided or required by the user, including documents, datasets, repositories,  tools, external agents, and existing agent graphs.

Traditional automated workflow design formulates the problem as search over a workflow space:
\vspace{-0.3em}
\begin{align}
    W^* = \arg\max_{W \in \mathcal{W}} \text{Eval}(W; D),
\end{align}
\vspace{-1.2em}

where \(\mathcal{W}\) is the candidate space, \(D\) is a benchmark or training set, and \(\text{Eval}\) is a scalar evaluation function. This formulation is effective when representative tasks and reliable scalar metrics exist, but many real-world and scientific tasks have no curated benchmark, no ground-truth output, and no single scalar reward that captures workflow quality. Success instead depends on heterogeneous evidence such as the validity of intermediate artifacts, correctness of tool use, scientific plausibility, and ability to recover from failures.

We therefore formulate automated workflow design as a \emph{retrieval-based synthesis} problem, in which a workflow is composed from retrieved resources, skills, tools, and external agents. 
Let $S$ denote a global library of reusable artifacts, partitioned into reference resources such as papers and documentation, procedural agent skills, callable tools, and wrapped external agent repositories. 
Each artifact carries a description and an I/O interface or typed artifact schema so that retrieved items can be composed. 
% We further denote an external agent registry \(\mathcal{A}\), kept separate from \(\mathcal{S}\) because external agents are runtime processes with their own state and interfaces rather than static artifacts. 
% AgentCo-op retrieves relevant items and synthesizes a workflow

Then \methodname synthesizes a workflow based on the retrieved artifacts:
\vspace{-0.5em}
\begin{align}
\label{eq:workflow_definition}
W = \textsc{Synthesize}(x, \mathcal{S}) \triangleq (R, G, \phi, \Pi),
\end{align}
where \(R\) is a set of agent roles, \(G\) is a dependency graph over roles, \(\phi: R \rightarrow 2^{\mathcal{S}}\) attaches a set of artifacts to each role, and \(\Pi\) specifies the interface protocol for communication between agents. 
\vspace{-0.5em}

\begin{figure}[!t]
    \centering
\includegraphics[width=1\linewidth]{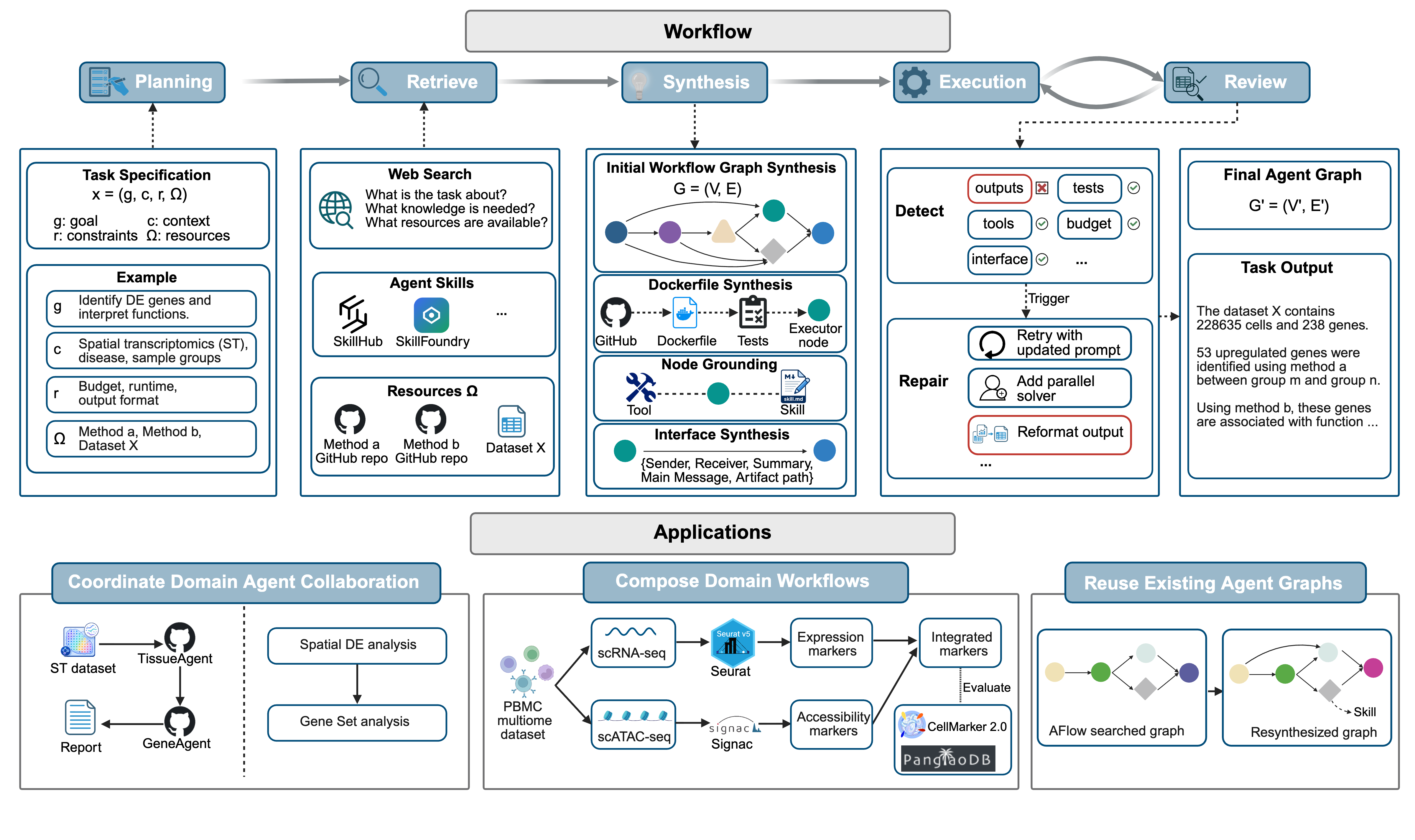}
    \vspace{-2.7em}
    \caption{Overview of \methodname.
\methodname synthesizes multi-agent workflows through five main stages: Planning, Retrieval, Synthesis, Execution, and Review. Given a typed task specification \(x=(g,c,r,\Omega)\), the system retrieves relevant knowledge, skills, tools, repositories, and datasets, then synthesizes an executable workflow graph \(G=(V,E)\). The synthesis stage includes initial graph construction, Dockerfile or executor wrapping, node grounding with skills and tools, and interface alignment through standardized message and artifact schemas. During execution, the reviewer monitors signals such as outputs, tests, tool behavior, budget, and interfaces. When failures or uncertainty arise, \methodname performs bounded local repair, producing a patched graph \(G'=(V',E')\) and the final task output. \methodname supports three representative applications: coordinating collaboration among domain-specific agents, composing domain workflows, and reusing existing agent graphs.
}
\vspace{-1.5em}
\label{fig:overview}
\end{figure}

\vspace{-0.5em}

\subsection{Method Details}
\vspace{-0.5em}
\paragraph{Graph representation and workflow synthesis.}
% \Cheng{Citation; }
% \Cheng{Cite GPTswarm for representing agents as graphs}
% \Cheng{Aflow explicitly said it used code as edge structure instead of DAG; ADAS also used code as the search sapce instead of graph}
% \Cheng{Define a multi-agent system as a grap G={V,E}...}
Following prior work such as GPTSwarm~\citep{zhuge2024gptswarm} and Flow~\citep{niu2025flow}, we represent a multi-agent workflow as a directed graph $G=(V,E)$ in which each node is an agent with an assigned role belongs to $R$, and each edge represents the direction in which information and intermediate artifacts flow between agents. 
The interface protocol is governed by $\Pi$ in Eq.~\ref{eq:workflow_definition}.
\methodname extends the graph representation in two ways. First, a node can be an external agent or an end-to-end method wrapped in a Docker container, so the graph can incorporate heterogeneous components without imposing a uniform native execution environment. 
Second, every agent node is equipped with a set of skills and tools matched to its role, and the mapping relationship is defined by $\phi$ in Eq.~\ref{eq:workflow_definition}. 
Thus, a node carries not only an instruction but also the procedural knowledge and callable operations needed to execute it.

To synthesize the workflow, \methodname analyzes the input task specification and formulates a retrieval plan. \methodname then retrieves a set of task-relevant artifacts, including related materials that inform the choice of workflow topology, agent skills that encode procedural knowledge, tools that expose callable operations, and metadata and documentation from the external GitHub repository if a GitHub repository URL is provided. \methodname analyzes these artifacts to synthesize an initial multi-agent workflow as a directed graph.
% sub-goals and dispatches multiple Retriever sub-agents in parallel to gather task-relevant artifacts from heterogeneous sources, including research papers, web pages, the SkillHub and SkillFoundry libraries, and tool registries. The retrieved artifacts both guide the construction of the workflow topology, by providing concrete references for how similar problems have been decomposed, and expand each agent node by attaching procedural knowledge and callable operations matched to its role.

\vspace{-0.5em}
\paragraph{Evidence-guided local repair.}
During workflow execution, \methodname continuously monitors execution evidence such as logs, intermediate outputs, validation signals, tool errors, and cost signals. 
A Reviewer triggers local repair when this evidence indicates failure or uncertainty.
Local repair consults a small set of repair policies and revises only the implicated nodes, attached skills and tools, or communication edges, so \methodname produces a patched graph \(G'=(V',E')\) rather than restarting the entire synthesis pipeline. 
Repair stops when validation succeeds, the repair budget is exhausted, or the maximum number of repair rounds is reached. 
This bounded, evidence-guided adaptation allows the workflow to recover from issues that emerge only at execution time and would be hard to anticipate during the initial synthesis.

\vspace{-1em}
\paragraph{Composition with domain workflows and search-based workflows.}
As shown in Fig.~\ref{fig:overview}, \methodname accepts a GitHub repository URL and wraps the external workflow into a Docker container. 
Following the approach of Repo2Run~\citep{hu2025repo2run}, \methodname builds the Docker image, uses build feedback and available tests to revise the container specification, and synthesizes or updates the Dockerfile together with accompanying documentation. 
The Docker container is then plugged into the graph as an external agent node, with its inputs and outputs aligned through the typed interface protocol. The same wrapping procedure applies to end-to-end methods, which can be attached to agent nodes as tools. Together, these mechanisms let \methodname reuse prior engineering, resolve environment dependency conflicts through Docker, and enable independently developed agents to collaborate on tasks that none of them could solve alone.

\methodname can also use an agent graph produced by a search-based agentic workflow design method as an additional resource in \(\Omega\). Rather than executing this graph directly, \methodname treats it as a reference structure that guides workflow synthesis. The framework then follows the same synthesis process illustrated in Fig.~\ref{fig:overview}, grounding graph nodes with retrieved skills and tools, enforcing typed interface and execution constraints, and applying bounded evidence-guided local repair during runtime.

\vspace{-1em}
\section{Experiments}
\vspace{-0.5em}
\label{sec:exp}
We evaluate \methodname in two complementary regimes. First, we study three open-world workflow composition tasks that motivate the synthesis-first design. 
The first task tests whether \methodname can coordinate independently developed domain agents by composing TissueAgent and GeneAgent~\citep{wang2025geneagent} for differential expression and gene-set interpretation on a developing human heart MERFISH dataset~\citep{farah2024spatially}. The second task examines parallel composition of complementary domain workflows, integrating Seurat~\citep{butler2018integrating} and Signac~\citep{stuart2021signac} into a cross-modality marker-discovery pipeline on PBMC multiome data, with marker quality evaluated against CellMarker 2.0~\citep{hu2023cellmarker} and PanglaoDB~\citep{franzen2019panglaodb}. The third task evaluates whether \methodname can reuse and improve a searched workflow by importing a multi-agent graph produced by AFlow~\citep{zhang2024aflow} on MBPP~\citep{austin2021mbpp}, then refining it through retrieval-based synthesis and bounded evidence-guided local repair.

We further evaluate \methodname on six standard benchmarks spanning question answering (HotpotQA~\citep{yang2018hotpotqa}, DROP~\citep{dua2019drop}), code generation (HumanEval~\citep{chen2021evaluating}, MBPP~\citep{austin2021mbpp}), and mathematical reasoning (GSM8K~\citep{cobbe2021gsm8k}, MATH~\citep{hendrycks2021math}). Following prior work~\citep{zhang2024aflow}, we use GPT-4o-mini as the base model for our method and matched-backbone baseline runs. 
We compare \methodname against three categories of baselines, including single-agent methods such as CoT~\citep{wei2022chain}, CoT SC~\citep{wang2022self}, Self Refine~\citep{madaan2023self}, and MedPrompt~\citep{nori2023can}, search-based multi-agent design methods such as ADAS~\citep{hu2024adas} and AFlow~\citep{zhang2024aflow}, and predefined multi-agent collaboration methods such as MultiPersona~\citep{wang2024unleashing}, LLM-Debate~\citep{du2024improving}, and Reconcile~\citep{chen2024reconcile}. Beyond task performance, we also analyze test-time and aggregate token cost of \methodname against multi-agent collaboration baselines to demonstrate its efficiency.

\begin{wrapfigure}{r}{0.4\textwidth}
\vspace{-1.8em}
\centering
    \includegraphics[width=\linewidth]{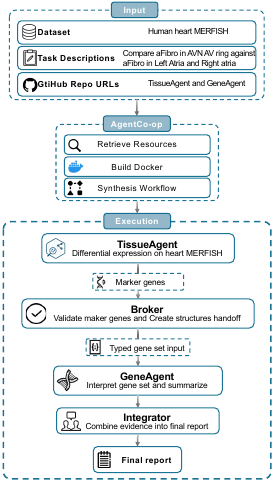}

\vspace{-0.5em}
\caption{\methodname orchestrates domain agents for collaborative biological analysis.
Given a developing human heart MERFISH dataset and a task description, \methodname prepares the domain-agent environment by profiling repositories and building containers, then coordinates a collaborative workflow for TissueAgent and GeneAgent.}
\label{fig:Domain_Agent_Collaboration}
\vspace{-2em}
\end{wrapfigure}

\vspace{-1.5em}
\subsection{Coordinate Domain Agent Collaboration}
\vspace{-0.5em}
\label{sec:Domain_Agent_Collaboration}

First, we evaluate whether \methodname can coordinate independently developed domain agents to solve a collaborative scientific analysis task. The task asks whether aFibro cells in the AVN/AV ring cellular community exhibit a distinct transcriptional program compared with aFibro cells in the left atria and right atria communities in a spatial transcriptomics dataset~\citep{farah2024spatially}. Solving this task requires spatial transcriptomics analysis to process the dataset and identify differentially expressed marker genes, followed by gene-set interpretation to characterize the resulting transcriptional program. The task naturally requires collaboration between a spatial transcriptomics agent and a gene-set analysis agent, making it a direct test of whether \methodname can synthesize a coherent workflow across independently developed scientific agents with different inputs, outputs, and execution environments.

We leverage \methodname to compose TissueAgent~\citep{tissueagent2025} and GeneAgent~\citep{wang2025geneagent}, which are specialized agents for spatial transcriptomics analysis and gene-set analysis, respectively. Given only the task description and the GitHub repository URLs for the two agents, \methodname profiles both repositories, builds isolated Docker containers, registers each container as an external workflow node, and synthesizes the collaborative workflow shown in Fig.~\ref{fig:composed_domain_workflows} with an explicit broker-mediated typed handoff between marker discovery and gene-set interpretation.

The synthesized workflow executes in four steps. First, the upstream TissueAgent node loads the MERFISH object, identifies 576 target aFibro cells in the AVN/AV ring community and 5,685 control aFibro cells in the left and right atria communities, and performs differential expression analysis, yielding 53 upregulated markers. Second, the Broker node validates the marker table and converts it into a structured input while preserving all 53 genes, ensuring that the downstream GeneAgent node receives typed evidence rather than an unstructured free-text list. 
Third, GeneAgent interprets the 53 markers without re-running differential expression and annotates them as an AV canal- and node-associated fibroblast program. Finally, the Integrator combines the differential expression evidence with the GeneAgent interpretation and concludes that AVN/AV ring aFibro cells represent a developmentally specialized, ECM-rich, conduction-niche-associated state rather than generic atrial stroma. This result demonstrates that \methodname can coordinate independently developed scientific agents into a coherent collaborative workflow to solve an end-to-end scientific analysis task without requiring global workflow search or redesigning either repository.

\vspace{-0.5em}
\subsection{Compose Domain Workflows}
\vspace{-0.5em}
\label{sec:pbmc_collaboration}
% We then demonstrate \methodname can compose existing domain workflows into a new collaborative scientific workflow. We use the PBMC Multiome dataset from 10X Genomics with paired RNA and ATAC measurements from the same cells. \methodname builds two separate Docker containers from the Seurat~\citep{butler2018integrating} and Signac~\citep{stuart2021single} Github Repositories, registers them as execution nodes, and synthesizes a parallel workflow followed by a join step as illustrated in Fig.~\ref{fig:composed_domain_workflows}. The Seurat node runs FindAllMarkers function on the gene expression assay, and the Signac node runs GeneActivity function followed by FindAllMarkers function on the gene activity scores. The evaluator collects identified marker sets from both nodes, computes their intersection and union per cell type, and computes them against CellMarker 2.0 and PanglaoDB\citep{hu2023cellmarker,franzen2019panglaodb}.
We then evaluate whether \methodname can compose independently developed domain workflows to solve a collaborative cross-modality analysis task. The task asks whether integrating marker signals from paired RNA and chromatin accessibility modalities can improve cell-type marker identification on multiome data. Solving this task requires scRNA-seq analysis to identify expression-based markers and scATAC-seq analysis to identify accessibility-based markers, since the two modalities provide partially overlapping but non-identical evidence of cell identity. It naturally requires coordination between RNA and ATAC analysis workflows. As such, this task provides a direct test of whether \methodname can synthesize a coherent cross-modality workflow from modality-specialized domain repositories and integrate complementary evidence across modalities.

\begin{figure}
    \centering
    \includegraphics[width=1\linewidth]{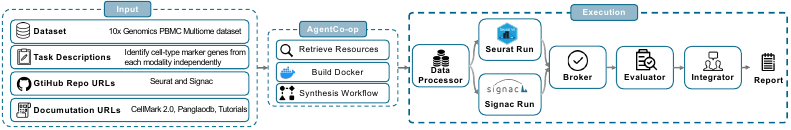}
    \vspace{-1.8em}
    \caption{\methodname coordinates external tools for cross-modal marker discovery. \methodname registers external Seurat and Signac tool nodes, runs parallel RNA and ATAC marker-discovery branches, validates typed artifacts, evaluates marker support against CellMarker 2.0 and PanglaoDB, and integrates the evidence into a final report.}
    \label{fig:composed_domain_workflows}
    \vspace{-1.5em}
\end{figure}

We leverage \methodname to compose Seurat~\citep{butler2018integrating} and Signac~\citep{stuart2021signac}, which are widely used workflows for single-cell RNA-seq and ATAC-seq analysis respectively. 
Given the task description, the PBMC multiome dataset from 10x Genomics, and the GitHub repositories and tutorials of the two workflows, \methodname builds two separate Docker containers from the Seurat and Signac GitHub repositories, registers them as execution nodes, and synthesizes a parallel workflow followed by a join step as illustrated in Fig.~\ref{fig:composed_domain_workflows}. 
The Seurat node runs the FindAllMarkers function on the gene expression assay, and the Signac node runs the GeneActivity function followed by the FindAllMarkers function on the chromatin accessibility assay. The evaluator collects identified marker sets from both nodes, computes their intersection and union per cell type, and evaluates them against CellMarker 2.0 and PanglaoDB~\citep{hu2023cellmarker,franzen2019panglaodb}, two established cell-type marker databases. The intersection captures jointly supported markers and is evaluated for precision, while the union captures all recovered markers and is evaluated for recall.

\begin{wraptable}{r}{0.46\textwidth}
\vspace{-\baselineskip}
\small
\centering
\vspace{-0.8em}
\caption{Macro precision and recall of cross-modality marker integration on the PBMC multiome dataset, evaluated against two marker gene databases. Precision is computed on the intersection and recall on the union of the two modalities. RNA and ATAC report each modality alone. Combined reports the cross-modality result. Bold marks the best result.}
\vspace{0.3em}
\label{tab:pbmc}
\setlength{\tabcolsep}{4pt}
\resizebox{\linewidth}{!}{%
\begin{tabular}{llccc}
\toprule
\textbf{Database} & \textbf{Metric} & \textbf{RNA} & \textbf{ATAC} & \textbf{Combined} \\
\midrule
\multirow{2}{*}{CellMarker 2.0} & Precision & 0.195 & 0.110 & \textbf{0.303} \\
                                & Recall    & 0.102 & 0.061 & \textbf{0.124} \\
\midrule
\multirow{2}{*}{PanglaoDB}      & Precision & 0.231 & 0.131 & \textbf{0.333} \\
                                & Recall    & 0.097 & 0.054 & \textbf{0.117} \\
\bottomrule
\end{tabular}%
}
\vspace{-\baselineskip}
\end{wraptable}

The results are shown in Tab.~\ref{tab:pbmc}. Across both reference databases, combining the two modalities improves both macro precision and recall over either single modality, and this trend holds at the per-cell-type level for the majority of cell types. This shows that \methodname can compose existing domain workflows to integrate complementary evidence. More details are illustrated in App.~\ref{app:parallel_collaboration}.

%maybe change to this
%This supports the claim that \methodname can automatically compose independently developed domain tools into a collaborative workflow that produces results superior to either tool alone.

\vspace{-0.5em}
\subsection{Reuse Existing Agent Graphs}
We further show that \methodname can import an existing predefined agent graph as a reference to synthesis the workflow. We take the multi-agent graph produced by a trained AFlow search and feed it into \methodname as additional resources in $\Omega$, then \methodname can retrieve the artifacts, resynthesize the agent graph and interface protocol, attach retrieved skills and tools to the agent nodes and apply bouneded local repair during execution. We evaluate the resulting workflow on the MBPP benchmark~\citep{austin2021mbpp}. The results are reported in Tab.~\ref{tab:imported}. AFlow + \methodname outperforms both AFlow alone and \methodname built from scratch, which confirms that synthesis and search are complementary. The imported graph contributes a strong reference resource, while \methodname contributes resource grounding and runtime adaptability that pure search does not provide.

\begin{wraptable}{r}{0.42\textwidth}
\vspace{-\baselineskip}
\small
\centering
\vspace{-1em}
\caption{MBPP performance of different agentic workflow design strategies. Bold indicates the best pass@1 score. Initializing \methodname from AFlow searched graph improves performance compared with initializing it from scratch.}
\vspace{0.8em}
\label{tab:imported}
\setlength{\tabcolsep}{6pt}
\resizebox{0.9\linewidth}{!}{%
\begin{tabular}{lc}
\toprule
\textbf{Strategy} & \textbf{pass@1} \\
\midrule
AFlow                       & 78.2 \\
\methodname (From Scratch)   & 87.1 \\
AFlow + \methodname          & \textbf{87.5} \\
\bottomrule
\end{tabular}%
}
\vspace{-0.5em}
\end{wraptable}

\vspace{-0.5em}
\subsection{Benchmarks}
\vspace{-0.3em}
\label{sec:bench}

The benchmark results are shown in Tab.~\ref{tab:benchmark_results}. 
For GSM8K and MATH, we report the Solve Rate (\%)as the primary metric.
For HumanEval and MBPP, we report the pass@1 metric to assess code accuracy. For HotpotQA and DROP, we report the F1 Score. 
Without any training or workflow-search stage, \methodname achieves the best performance on four of the six benchmarks and ranks first on the average score under the matched backbone setting.
% We believe \methodname can match search-based methods because it does not search blindly over workflow space. 
This competitiveness is consistent with \textsc{AgentCo-op}’s synthesis-first design: 
It starts from reusable workflow priors, including skills, roles, motifs, typed handoffs, and repair policies, which already encode many of the structures search methods have to discover through expensive trial and error. 
% And search methods usually optimize one persistent workflow for a task distribution. But different benchmark instances often need different fixes.
Search-based methods typically optimize one persistent workflow for a task distribution, whereas different benchmark instances often require different checks, decompositions, or repair actions. 
Instead of searching for one workflow that works best on average, \methodname adapts the workflow locally for each problem when execution evidence indicates failure or uncertainty.

\begin{table}[!h]
\centering
\vspace{-0.5em}
\caption{Performance comparison of different methods on six benchmarks spanning QA, code, and math, using GPT-4o-mini as the backbone model. Bold indicates the best result. The original AFlow paper mixes multiple backbone models, and AFlow* denotes the results reported by its authors. For a fair comparison, we rerun AFlow using GPT-4o-mini only, reported as AFlow (GPT-4o-mini).}
\label{tab:benchmark_results}
\vspace{-0.5em}
\resizebox{\linewidth}{!}{%
\begin{tabular}{lccccccc}
\toprule
\multirow{2}{*}{\textbf{Method}} & \multicolumn{6}{c}{\textbf{Benchmarks}} & \multirow{2}{*}{\textbf{Avg.}} \\
\cmidrule(lr){2-7}
 & HotpotQA & DROP & HumanEval & MBPP & GSM8K & MATH & \\
\midrule
IO (GPT-4o-mini)~\citep{hurst2024gpt}                       & 68.1 & 68.3          & 87.0          & 71.8 & 92.7 & 48.6 & 72.8 \\
CoT~\cite{wei2022chain}                 & 67.9 & 78.5          & 88.6          & 71.8 & 92.4 & 48.8 & 74.7 \\
CoT SC (5-shot)~\cite{wang2022self}     & 68.9 & 78.8          & 91.6          & 73.6 & 92.7 & 50.4 & 76.0 \\
MedPrompt~\cite{nori2023can}            & 68.3 & 78.0          & 91.6          & 73.6 & 90.0 & 50.0 & 75.3 \\
MultiPersona~\cite{wang2024unleashing}  & 69.2 & 74.4          & 89.3          & 73.6 & 92.8 & 50.8 & 75.0 \\
Self Refine~\cite{madaan2023self}       & 60.8 & 70.2          & 87.8          & 69.8 & 89.6 & 46.1 & 70.7 \\
ADAS~\cite{hu2024adas}             & 64.5 & 76.6          & 82.4          & 53.4 & 90.8 & 35.4 & 67.2 \\
AFlow*~\cite{zhang2024aflow}                                  & 73.5 & 80.6          & \textbf{94.7} & 83.4 & 93.5 & 56.2 & 80.3 \\
\midrule
LLM-Debate~\cite{du2024improving}       & 71.8 & 81.4          & 91.4          & 70.7 & 92.4 & 50.0 & 76.3 \\
ReConcile~\cite{chen2024reconcile}   & 73.8 & \textbf{82.1} & 89.3          & 70.3 & 93.7 & 44.1 & 75.6 \\
AFlow (GPT-4o-mini)~\cite{zhang2024aflow}                  & 71.4 & 68.9          & 89.3          & 78.2 & 86.8 & 53.1 & 74.3 \\
\rowcolor{lm_purple}\methodname (GPT-4o-mini)                & \textbf{76.5} & 77.2 & 90.2 & \textbf{87.1} & \textbf{94.4} & \textbf{58.2} & \textbf{80.6} \\
\bottomrule
\end{tabular}%
}
\vspace{-0.5em}

\end{table}
 
\vspace{-0.5em}
\subsection{Cost Analysis}
\vspace{-0.5em}
We further record the token cost of each method on every benchmark, with results shown in Tab.~\ref{tab:cost_analysis}. 
\methodname is substantially more efficient than the multi-agent baselines, and its 
test-time cost is lower than ReConcile on all six benchmarks and lower than LLM Debate on five of six benchmarks. 
%per-task cost is lower on most benchmarks. 
Search-based methods such as AFlow consume additional tokens and time exploring, evaluating, and optimizing candidate workflows on training set before producing a final design. 
Discussion-based methods such as LLM-Debate and ReConcile incur repeated rounds of inter-agent communication for every task instance, which compounds quickly as the number of tasks grows. 
\methodname avoids both of these patterns. 
\methodname separates workflow synthesis from workflow repair. Synthesis produces a reusable initial workflow, while repair performs bounded, instance-specific local repair during execution time.
\methodname modifies the workflow graph only for the current instance and the modifications will be discarded afterward. 
This design allows \methodname to adapt to heterogeneous problem instances without overfitting the global workflow or requiring expensive resynthesis.

\begin{table}[t]
\small
\centering
\caption{Per-dataset performance and cost comparison across methods. Costs are aggregated over the entire benchmark dataset. A dash in the Train Cost column indicates that the method requires no workflow search or training stage. }
\vspace{-0.3em}
\label{tab:cost_analysis}
\resizebox{0.65\linewidth}{!}{%
\begin{tabular}{llrrrr}
\toprule
\textbf{Dataset} & \textbf{Method} & \textbf{Score} & \textbf{Train Cost} & \textbf{Test Cost} & \textbf{Total Cost} \\
\midrule
\multirow{4}{*}{HotpotQA}
 & LLM Debate      & 71.8          & -- & \$1.5200 & \$1.5200 \\
 & ReConcile       & 73.8          & -- & \$3.7600 & \$3.7600 \\
 & AFlow           & 20.0          & \$4.6104 & \$1.3398 & \$5.9502 \\
 \rowcolor{lm_purple}& \methodname  & \textbf{76.5} & -- & \$0.4284 & \$0.4284 \\
\midrule
\multirow{4}{*}{DROP}
 & LLM Debate      & 81.4          & -- & \$0.7200 & \$0.7200 \\
 & ReConcile       & \textbf{82.1} & -- & \$1.6800 & \$1.6800 \\
 & AFlow           & 68.9          & \$1.6798 & \$0.3235 & \$2.0033 \\
 \rowcolor{lm_purple}& \methodname  & 77.2          & -- & \$0.3853 & \$0.3853 \\
\midrule
\multirow{4}{*}{HumanEval}
 & LLM Debate      & \textbf{91.4} & -- & \$0.1572 & \$0.1572 \\
 & ReConcile       & 89.3          & -- & \$0.4061 & \$0.4061 \\
 & AFlow           & 89.3          & \$0.2258 & \$0.0371 & \$0.2629 \\
 \rowcolor{lm_purple}& \methodname  & 90.2          & -- & \$0.1062 & \$0.1062 \\
\midrule
\multirow{4}{*}{MBPP}
 & LLM Debate      & 70.7          & -- & \$0.1705 & \$0.1705 \\
 & ReConcile       & 70.3          & -- & \$0.7502 & \$0.7502 \\
 & AFlow           & 72.4          & \$0.3475 & \$0.1152 & \$0.4627 \\
 \rowcolor{lm_purple}& \methodname  & \textbf{87.1} & -- & \$0.1791 & \$0.1791 \\
\midrule
\multirow{4}{*}{GSM8K}
 & LLM Debate      & 92.4          & -- & \$1.6880 & \$1.6880 \\
 & ReConcile       & 93.7          & -- & \$1.8990 & \$1.8990 \\
 & AFlow           & 86.8          & \$0.0469 & \$0.2000 & \$0.2469 \\
 \rowcolor{lm_purple}& \methodname  & \textbf{94.4} & -- & \$0.2537 & \$0.2537 \\
\midrule
\multirow{4}{*}{MATH}
 & LLM Debate      & 50.0          & -- & \$1.7982 & \$1.7982 \\
 & ReConcile       & 44.1          & -- & \$1.6038 & \$1.6038 \\
 & AFlow           & 53.1          & \$0.0781 & \$0.2691 & \$0.3472 \\
 \rowcolor{lm_purple}& \methodname  & \textbf{58.2} & -- & \$0.3670 & \$0.3670 \\
\bottomrule
\end{tabular}%
}
\vspace{-1.5em}

\end{table}

% For each benchmark, it retrieves a set of reusable artifacts once and synthesizes the agent graph directly from them, without any search loop. For each individual task, the synthesized graph is then executed as is, and tokens are spent on additional reasoning only when local repair is triggered. This design preserves the strong accuracy reported in the previous section while reducing token cost by a large margin.
\vspace{-0.5em}
\section{Limitations and Future Work}
\vspace{-0.5em}
We acknowledge several limitations of this study. 
First, our current evaluation of domain-agent collaboration remains within a single scientific domain and two genomics-centered case studies. 
Future work could extend this setting to interdisciplinary synthesis and cross-domain agent collaboration. Second, the framework still depends on the quality of the available domain resources, including specialized agents, skills, and tools. Coordination failures may occur when agents produce poorly specified outputs, expose incompatible interfaces, or generate intermediate artifacts that are difficult to validate.
Third, bounded local repair improves robustness but does not guarantee global optimality; a locally repaired workflow may still miss a better global organization. 
Fourth, the biological case studies demonstrate auditable workflow composition, but their scientific conclusions should be interpreted as computational analyses that require expert review and, where appropriate, orthogonal validation. 
In future work, we plan to extend \methodname toward more adaptive organization discovery, stronger verification of intermediate outputs, richer memory and provenance tracking, more explicit typed artifact schemas, and broader integration with domain-specific agent libraries. 

\vspace{-0.8em}
\section{Conclusion}
\vspace{-0.8em}

In summary, \methodname introduces a retrieval-based synthesis paradigm for automatic multi-agent workflow design. 
Rather than relying on benchmark-driven global search, \methodname dynamically composes skills, tools, roles, and external agents into task-specific workflows, coordinates them through typed artifact handoffs, and refines implicated components through bounded evidence-guided local repair. 
Our results show that workflow synthesis can remain competitive with search-based design methods on standard benchmarks while reducing test-time token cost relative to discussion-based multi-agent baselines. 
More importantly, this paradigm is better aligned with real-world scientific domains, where benchmarks are often inaccessible, prior workflows already exist, and interoperability among heterogeneous agents is essential. 
In two biological analysis case studies, \methodname demonstrates how independently developed agents and domain methods can be coordinated to answer biologically meaningful questions that require multiple complementary forms of expertise without redesigning the underlying repositories or performing global topology search.

More broadly, this work suggests that progress in scientific agents will depend not only on building stronger specialized agents, but also on developing methods for organizing, composing, and adapting heterogeneous agents into reliable collaborative systems. 
Scientific workflows are rarely solved by a single capability; they require coordinated expertise across multiple tasks and domains. 
By treating workflow construction as an organization synthesis problem, \methodname provides a practical path toward a reusable, cooperative, and extensible ecosystem of specialized scientific agents, where existing expertise can be composed, verified, and reused to address increasingly complex cross-domain discovery challenges.

% \section*{References}

% References follow the acknowledgments in the camera-ready paper. Use unnumbered first-level heading for
% the references. Any choice of citation style is acceptable as long as you are
% consistent. It is permissible to reduce the font size to \verb+small+ (9 point)
% when listing the references.
% Note that the Reference section does not count towards the page limit.
% \medskip

% {
% \small

% [1] Alexander, J.A.\ \& Mozer, M.C.\ (1995) Template-based algorithms for
% connectionist rule extraction. In G.\ Tesauro, D.S.\ Touretzky and T.K.\ Leen
% (eds.), {\it Advances in Neural Information Processing Systems 7},
% pp.\ 609--616. Cambridge, MA: MIT Press.

% [2] Bower, J.M.\ \& Beeman, D.\ (1995) {\it The Book of GENESIS: Exploring
%   Realistic Neural Models with the GEneral NEural SImulation System.}  New York:
% TELOS/Springer--Verlag.

% [3] Hasselmo, M.E., Schnell, E.\ \& Barkai, E.\ (1995) Dynamics of learning and
% recall at excitatory recurrent synapses and cholinergic modulation in rat
% hippocampal region CA3. {\it Journal of Neuroscience} {\bf 15}(7):5249-5262.
% }

%%%%%%%%%%%%%%%%%%%%%%%%%%%%%%%%%%%%%%%%%%%%%%%%%%%%%%%%%%%%

\newpage

\appendix

\bibliographystyle{plainnat}

\bibliography{refs}

%%%%%%%%%%%%%%%%%%%%%%%%%%%%%%%%%%%%%%%%%%%%%%%%%%%%%%%%%%%%

\clearpage
\section{Appendix}

\subsection{Methods}
\label{app:method}

\subsubsection{Planning and Retrieval}
\label{app:method-plan-retrieve}

\paragraph{Planning.}
The Planning stage analyzes the typed task specification $x = (g, c, r, \Omega)$ and formulates a retrieval plan that determines what knowledge is needed to solve the task. Concretely, \methodname decomposes the user goal $g$ into sub-goals consistent with the task context $c$, identifies the operational constraints in $r$ that must be respected during synthesis, including runtime, budget, environment requirements, and desired output format, and inspects the resources in $\Omega$ to decide which entries should be wrapped as execution nodes and which can be referenced as documents or datasets. The retrieval plan therefore acts as a specification of what \methodname should look up before constructing the graph, separating the decision of \emph{what to retrieve} from the actual retrieval calls.

\paragraph{Retrieval.}
Guided by the retrieval plan, \methodname gathers task-relevant artifacts from heterogeneous sources. Reference resources, including research papers and curated documentation, inform the choice of workflow topology by providing concrete examples of how similar problems have been decomposed. Agent skill libraries, such as SkillHub and SkillFoundry, supply procedural knowledge encoded as portable skill packages. Tool registries expose callable operations such as database queries, plotting routines, or domain-specific functions. For each external repository URL provided in $\Omega$, \methodname additionally retrieves the repository metadata and documentation, including README files, tutorials, and example scripts, which are later used both for synthesizing the Docker container and for grounding the corresponding executor node. The retrieved artifacts populate the global library $\mathcal{S}$ from which the workflow is later composed.

\subsubsection{Synthesis}
\label{app:method-synthesis-detail}

\paragraph{Initial graph construction.}
The Synthesis stage begins by producing an initial directed graph $G = (V, E)$ from the retrieved artifacts and task specification. Topology decisions are informed by retrieved reference workflows for related problems and by the structure of any imported agent graph in $\Omega$. \methodname also decides whether the graph should be linear, parallel, or a mixed topology based on data dependencies in the task, for example assigning two modality-specific analyses to parallel branches when their inputs are independent and joining them at a downstream evaluator.

\paragraph{Node grounding.}
After the topology is fixed, \methodname grounds each node by attaching a set of skills and tools matched to its role, defined by the mapping $\phi : R \rightarrow 2^{\mathcal{S}}$ in Eq.~\ref{eq:workflow_definition}. Matching is performed by scoring candidate skills and tools against the role description and the upstream and downstream artifact types of the node, and selecting the top-ranked entries. As a result, every node carries not only an instruction but also the procedural knowledge and callable operations needed to execute it, which both reduces prompt-engineering load and enforces consistency across instances of the same role.

\paragraph{Dockerfile synthesis.}
For each external repository or end-to-end method that needs to be wrapped as an execution node, \methodname builds an isolated Docker container following Repo2Run~\citep{hu2025repo2run}. The procedure is iterative: \methodname drafts a Dockerfile from the retrieved repository metadata, attempts to build the image, and on failure inspects the build log to revise the dependency list, base image, or build commands. Available repository tests and example scripts are then executed inside the container as a smoke check; recurrent failures trigger a further revision round. The same wrapping procedure also applies to end-to-end methods that are not full agentic workflows. Such methods are still packaged into Docker containers, but \methodname attaches them to existing agent nodes as callable tools rather than instantiating them as standalone executor nodes, which avoids unnecessary inter-node communication for components that only expose a single entry point.

\paragraph{Interface synthesis.}
Finally, \methodname synthesizes the interface protocol $\Pi$ that governs communication along each edge of $G$. Every communication step is described by a structured message that records the sender, the receiver, a short summary, the main message body, and the path of any typed artifact passed between the two nodes. Typed artifacts include validated marker tables, structured gene-set inputs, intermediate code files, and tool outputs serialized as JSON. The schema is enforced by the Broker nodes shown in Fig.~\ref{fig:overview}, which validate that the artifact produced by an upstream node satisfies the expected schema before it is consumed downstream. This explicit schema is what allows independently developed components, including Docker-wrapped repositories, to exchange information through validated artifacts rather than free-form text.

\subsubsection{Review Loop}
\label{app:method-reviewer}

\paragraph{Detect.}
\methodname aggregates execution evidence into a small set of structured signals. Output signals capture node-level results and judge confidences, test signals capture pass and fail counts on validation cases, tool signals capture tool invocation errors and missing outputs, budget signals capture accumulated token cost relative to the per-task budget in $r$, and interface signals capture schema mismatches between artifacts produced by an upstream node and the schema expected by a downstream node. \methodname flags a node as failing or uncertain when one or more of these signals cross a policy-specific threshold.

\paragraph{Decide.}
\methodname then matches the observed evidence pattern against a small set of repair policies. Each policy maps an evidence pattern to a repair action. Examples include retrying with an updated prompt when a judge node returns low confidence, adding a parallel solver when a code node has persistent test failures, swapping the backend when tool errors recur on the same external service, and reformatting the output of an upstream node when a downstream artifact violates its schema. Policies are evaluated in priority order, and the first matching policy determines the action passed to the Repair step.

\subsection{Experiments}

\subsubsection{Ablation Study}
\label{sec:ablation}
To verify the contribution of each component in \methodname, we conduct an ablation study with results reported in Tab.~\ref{tab:ablation}. After removing the runtime local repair, most benchmarks show a drop in accuracy. The remaining two benchmarks fluctuate within a small margin, which suggests that local repair contributes most when tasks involve longer reasoning chains or precise generation. When we further remove agent skills and tools, the performance on most benchmarks remains close to the \methodname without local repair. We attribute this to the nature of the standard benchmarks used here, which mainly require general reasoning and coding ability rather than specialized procedural knowledge or external operations. Agent skills and tools therefore have limited room to improve performance in this setting, although they are essential in the open-ended scientific scenarios.

\begin{table}[ht]
\small
\centering
\caption{Ablation study on the components of \methodname. \textbf{AC-Full} is the complete \methodname system; \textbf{AC-NoLocalRepair} removes runtime gates and local repair; \textbf{AC-Minimal} removes agent skills, tools, runtime gates and local repair. All values are accuracy (\%); bold marks the best result in each column.}
\label{tab:ablation}
\resizebox{0.8\linewidth}{!}{%
\begin{tabular}{lccccccc}
\toprule
\multirow{2}{*}{\textbf{Variant}} & \multicolumn{6}{c}{\textbf{Benchmarks}} & \multirow{2}{*}{\textbf{Avg.}} \\
\cmidrule(lr){2-7}
 & HotpotQA & DROP & HumanEval & MBPP & GSM8K & MATH & \\
\midrule
AC-Full              & 76.5          & 77.2          & \textbf{90.2} & \textbf{87.1} & \textbf{94.4} & \textbf{58.2} & \textbf{80.6} \\
AC-NoLocalRepair            & \textbf{76.6} & 77.4          & 87.9          & 86.8          & 93.2          & 56.6          & 79.8 \\
AC-Minimal           & 76.0          & 77.0          & 88.6          & 86.0          & 93.9          & 51.7          & 78.9 \\
\bottomrule
\end{tabular}%
}
\end{table}

\subsubsection{Coordinate Domain Agent Collaboration}
\label{app:serial_collaboration}
For this task, we evaluate whether \methodname can coordinate independently developed domain agents to solve a collaborative scientific analysis problem. The biological question asks whether aFibro cells in the AVN/AV ring cellular community exhibit a distinct transcriptional program compared with aFibro cells in the left atria and right atria communities in a developing human heart MERFISH dataset~\citep{farah2024spatially}. Answering the question requires more than generic cell-type comparison: the workflow must identify spatially localized differential expression signals and then interpret the resulting marker genes to characterize the resulting transcriptional program.

This setting requires \methodname to synthesize a collaborative workflow that composes two independently developed agents: TissueAgent and GeneAgent. TissueAgent is a role-based multi-agent framework that turns open-ended natural-language spatial transcriptomics requests and multimodal inputs into auditable, runnable workflows~\citep{tissueagent2025}. GeneAgent is an agent for gene-set analysis that reduces hallucinations by autonomously interacting with biological databases to verify its own outputs~\citep{wang2025geneagent}. Together, these agents provide complementary expertise, but they were developed independently and expose different repositories, interfaces, execution environments, and output formats.

Therefore, this task is also significant from an agent-composition perspective. No single specialized agent is sufficient on its own: spatial transcriptomics analysis is needed to load the MERFISH data, select the relevant cellular communities, and perform differential expression, while gene-set interpretation is needed to determine whether the marker genes correspond to a meaningful biological program. The task naturally requires collaboration between a spatial transcriptomics agent and a gene-set analysis agent, making it a direct test of whether \methodname can synthesize a coherent workflow across independently developed scientific agents with different inputs, outputs, and execution environments.

Given the task description, the two external GitHub repositories, and the public MERFISH dataset, \methodname compiles a serial collaborative workflow. The synthesized workflow consists of repository profiling, sandbox construction, agent registration, TissueAgent execution, broker validation, GeneAgent execution, integration, and reporting. The workflow is linear because GeneAgent depends on the marker-gene artifact produced by the upstream differential expression analysis. To bridge this interface, the Broker validates the TissueAgent marker table and converts it into a structured JSON input containing all 53 human genes, with zero genes dropped. As a result, GeneAgent receives the marker set as a typed artifact rather than an unstructured free-text list.

The upstream execution loaded the MERFISH AnnData object without synthetic fallback. The object contained 228635 cells and 238 genes. The detected annotation fields were \texttt{populations}, \texttt{communities}, and \texttt{sample\_id}. The target group was aFibro cells in the AVN/AV ring community. The control group was aFibro cells in the Left Atria and Right Atria communities. The target group contained 576 cells. The control group contained 5685 cells. The target cells were 295 in R78\_4C12 and 281 in R77\_4C4. The control cells were 2378 in R77\_4C4, 1849 in R78\_4C12, and 1458 in R78\_4C15.

\begin{table}[t]
\centering
\small
\caption{Summary of the serial collaboration case study.}
\label{tab:serial_collaboration_summary}
\begin{tabular}{p{0.30\linewidth}p{0.64\linewidth}}
\toprule
Item & Result \\
\midrule
External repositories & TissueAgent and GeneAgent \\
Workflow type & Linear handoff through a validated marker gene artifact \\
Dataset & Developing human heart MERFISH AnnData object \\
Dataset size & 228635 cells and 238 genes \\
Target group & 576 aFibro cells in the AVN/AV ring community \\
Control group & 5685 aFibro cells in the Left Atria and Right Atria communities \\
Primary differential expression & Welch testing with Benjamini Hochberg correction \\
Primary marker rule & Adjusted \(P < 0.05\) and positive \(\log_2\) fold change \\
Primary marker count & 53 AVN/AV ring upregulated markers \\
Sensitivity marker count & 46 markers with a Mann Whitney test \\
Sanity check genes & DES, HAND2, IGFBP5, MYH6, MYH7, and NELL2 were all recovered \\
Broker output & Structured GeneAgent input with 53 genes and zero dropped genes \\
GeneAgent label & AV canal and node associated fibroblast program \\
Final interpretation & AVN/AV ring aFibro cells support a specialized developmental and conduction associated state rather than generic atrial stroma \\
\bottomrule
\end{tabular}
\end{table}

The primary analysis normalized expression to a total count of 10000 per cell and then applied log transformation. Differential expression used Welch testing across the 238 MERFISH genes. Multiple testing correction used Benjamini Hochberg adjustment. The resulting marker list began with DES, MYH6, IGFBP5, MYH7, HAND2, HCN4, TBX3, NELL2, COL9A2, and CD34. A sensitivity run used the Mann Whitney test and returned 46 markers. We report the Welch result as the primary analysis because it was the executed primary configuration. We report the Mann Whitney result as a sensitivity analysis because it shows that the exact marker count changes with the test choice.

\begin{table}[t]
\centering
\small
\caption{GeneAgent interpretation of the validated marker genes.}
\label{tab:serial_geneagent_subprocesses}
\begin{tabular}{p{0.35\linewidth}p{0.59\linewidth}}
\toprule
Subprocess & Supporting genes reported by GeneAgent \\
\midrule
AV canal and conduction system specification & TBX3, TBX5, NKX2-5, HAND2, HCN4, HEY1, IRX4, SEMA6D \\
Extracellular matrix synthesis and remodeling & POSTN, FBLN2, FMOD, COL9A2, MMP11, CTSV, IGFBP5, TPBG \\
Epicardial and mesenchymal lineage identity & TCF21, WT1, PDGFRA, PRRX1, CD34, MECOM, TSHZ2 \\
Contractile and ion channel signals & MYH6, MYH7, TTN, DES, CACNA1C, KCNH2, RRAD, CNN1 \\
Neurogenic and axon guidance cues & NELL2, NRXN1, NEFL, NTS, PENK, SERPINI1, ADGRL1, BRINP3, SLC1A3, ADM \\
Developmental morphogen patterning & BMP2, INHBA, RSPO3, SFRP1, HHIP, BAMBI, CRABP2, MSX2 \\
\bottomrule
\end{tabular}
\end{table}

GeneAgent returned Markdown and JSON reports. Its self verification field listed Gene Ontology Biological Process, Reactome Pathways, UniProt and NCBI Gene summaries, Human Protein Atlas, and literature evidence on AV canal and AV node development. The final integrator combined the differential expression evidence and the GeneAgent report. It also retained caveats about cardiomyocyte-like transcripts, possible spatial admixture, and the need for orthogonal validation. These caveats are important because genes such as MYH6, MYH7, TTN, and DES may reflect proximity to nodal or transitional myocardium. The main systems conclusion is not affected by this caveat. \methodname converted two external repositories into coordinated execution nodes and preserved an auditable handoff from marker discovery to gene set interpretation.

\subsubsection{Compose Domain Workflows}
\label{app:parallel_collaboration}

For this task, we evaluate whether \methodname can compose existing domain workflows to solve a collaborative cross-modality analysis problem. The biological question asks whether integrating cell-type marker signals from paired RNA and chromatin accessibility modalities can yield more reliable cell-type marker identification than either modality alone on a single-cell multiome dataset. Answering this question requires more than running a single analysis pipeline. The workflow needs to independently identify markers from gene expression and from chromatin accessibility, and then reconcile their outputs against curated cell-type marker references.

This setting requires \methodname to synthesize a collaborative workflow that composes two domain workflows, Seurat and Signac. Seurat is a widely used analysis framework for single-cell RNA-seq that supports normalization, clustering, and differential expression on gene expression assays~\citep{butler2018integrating}. Signac extends single-cell analysis to chromatin accessibility, providing modality-specific quality control and a GeneActivity function that summarizes accessibility into per-gene scores~\citep{stuart2021signac}. Together, the two packages provide complementary expertise on the two modalities, but they are maintained as separate repositories with distinct dependencies, and integrating them into an automated workflow typically requires manual scripting and environment management.

Therefore, this task offers a meaningful workflow-composition perspective. The two modality-specific workflows can be executed in parallel since neither workflow consumes the other's output, while their results still need to be reconciled at the cell-type level to evaluate the integrated marker set. The task therefore naturally requires parallel coordination of two domain workflows followed by a join step, making it a direct test of whether \methodname can synthesize a coherent cross-modality workflow across existing domain repositories with different inputs and execution environments.

\begin{table}[h]
\centering
\small
\caption{PBMC workflow and marker discovery summary.}
\label{tab:pbmc_lite_workflow_summary}
\begin{tabular}{ll}
\toprule
Quantity & Value \\
\midrule
Dataset & 10x Genomic multiome dataset for human PBMCs \\
Assays & Gene expression and chromatin accessibility \\
Raw cells & 11909 \\
Post-QC cells & 11070 \\
Annotated cells before marker filter & 5000 \\
Cells used for marker discovery & 4777 \\
Cell types used for marker discovery & 22 \\
RNA marker rows & 26353 \\
ATAC gene activity marker rows & 45731 \\
Top markers per modality and cell type & 50 \\
Mean RNA and ATAC intersection size & 11.55 \\
Mean RNA and ATAC union size & 88.45 \\
Mean Jaccard index & 0.133 \\
\bottomrule
\end{tabular}
\end{table}

The inputs to the task include the Seurat and Signac GitHub repositories and official tutorials, the 10x PBMC multiome data files, the Hao PBMC reference data ~\citep{hao2021integrated}, and two marker databases, CellMarker 2.0 and PanglaoDB. \methodname compiles a parallel-then-join workflow with repository profiling, sandbox preparation, agent registration, data inspection, independent Seurat and Signac execution, broker validation, marker set evaluation, integration, and reporting. The broker records two valid typed handoffs, so the cross-modality result is evaluated through structured artifacts rather than free text.

\begin{table}[ht]
\centering
\small
\caption{Macro precision and recall results for the PBMC case study.}
\label{tab:pbmc_lite_precision_recall}
\begin{tabular}{lccccccc}
\toprule
Database & Cell types & \(P_{\cap}\) & \(P_{\mathrm{RNA}}\) & \(P_{\mathrm{ATAC}}\) & \(R_{\cup}\) & \(R_{\mathrm{RNA}}\) & \(R_{\mathrm{ATAC}}\) \\
\midrule
CellMarker 2.0 & 22 & 0.303 & 0.195 & 0.110 & 0.124 & 0.102 & 0.061 \\
PanglaoDB & 22 & 0.333 & 0.231 & 0.131 & 0.117 & 0.097 & 0.054 \\
\bottomrule
\end{tabular}
\end{table}

The intended annotation path transfers \texttt{celltype.l2} labels from the Hao reference. This path does not complete on the execution host, triggering \methodname's self-correction, which switches to a SingleR-based fallback with the MonacoImmuneData reference. The fallback produces 5000 annotated cells with 27 fine immune labels. After applying a minimum of 30 cells per label to ensure stable marker discovery, 4777 cells across 22 labels are retained for marker discovery. As a result, the quantitative evaluation reflects Monaco-derived immune labels rather than Hao \texttt{celltype.l2} labels. A summary of the workflow and marker discovery statistics is provided in Tab.~\ref{tab:pbmc_lite_workflow_summary}.

% The PBMC task supplied the Seurat and Signac GitHub repositories, the 10x PBMC multiome data files, the Hao PBMC reference~\citep{hao2023dictionary}, and two marker databases. \methodname compiled a parallel-then-join workflow with repository profiling, sandbox preparation, agent registration, data inspection, independent Seurat and Signac execution, broker validation, marker set evaluation, integration, and reporting. The broker recorded two valid typed handoffs, so the cross-modality result was evaluated through structured artifacts rather than free text.
 
% The intended annotation path was a Hao \texttt{celltype.l2} reference transfer. That path did not complete on the execution host, so the run used the documented SingleR fallback with the MonacoImmuneData reference. This produced 5000 annotated cells with 27 fine immune labels. After applying a minimum of 30 cells per label, 4777 cells across 22 labels were retained for marker discovery. As a result, the quantitative evaluation reflects Monaco-derived immune labels rather than Hao \texttt{celltype.l2} labels.

For each evaluated cell type, the evaluator constructs four marker sets, namely RNA markers, ATAC gene activity markers, their intersection, and their union. For a predicted marker set \(M\) and a database marker set \(D\), precision is computed as \(|M \cap D| / |M|\) and recall as \(|M \cap D| / |D|\). The precision comparison uses the intersection, RNA, and ATAC marker sets, and the recall comparison uses the union, RNA, and ATAC marker sets. The macro-averaged scores across cell types are summarized in Tab.~\ref{tab:pbmc_lite_precision_recall}. At the per-cell-type level, on CellMarker 2.0, the intersection improves precision over both single modalities for 17 of 22 cell types, and the union improves recall over both single modalities for 16 of 22 cell types. On PanglaoDB, the intersection improves precision for 15 of 22 cell types, the union improves recall for 18 of 22 cell types. \methodname thus composes two specialized domain repositories into coordinated parallel execution nodes and aligns their cross-modality marker outputs.

% These results support the main collaboration hypothesis at the macro level. The intersection of RNA and ATAC gene activity markers is more precise because it retains genes supported by both molecular modalities, and the union has higher recall because it keeps complementary genes from either modality. The result should not be interpreted as a universal per-cell-type guarantee, since several cell types deviate from the full ordering. The current artifact package reports the primary CellMarker 2.0 and PanglaoDB evaluations and does not contain the combined-database mode or the full sensitivity grid, which are therefore not reported here.

% \input{checklist.tex}

\end{document}